%% file: root.tex
\documentclass[dvipsnames]{article}

\usepackage[nonatbib, final]{corl_2020} %

\usepackage[export]{adjustbox}

\usepackage[basicpaper,fig,lengthhacks]{tinyiu}
\usepackage{xcolor}
\usepackage[font=scriptsize]{caption}

\usepackage[title]{appendix}

\usepackage{xifthen}
\usepackage{tikz}
\usepackage{wrapfig}
\usepackage{minibox}

\DeclareCaptionStyle{figstyle}
[margin=0pt,justification=centering]
{format=hang,margin=15pt,
    font=scriptsize}
\usepackage{floatrow}
\newfloatcommand{capbtabbox}{table}[][\FBwidth]

\input{_env_vars.tex}

\usepackage[boxed,ruled,vlined,linesnumbered]{algorithm2e}
\DontPrintSemicolon
\SetKwProg{Fn}{function}{}{}
\SetKwFunction{FnSampleFree}{SampleFree}
\SetKwFunction{FnRestartArm}{RestartArm}
\SetKwFunction{FnPickArm}{PickArm}
\SetKwFunction{FnRewire}{Rewire}
\SetKwFunction{FnNearest}{Nearest}
\SetKwFunction{FnRestartArm}{RestartArm}
\SetKwComment{Comment}{$\triangleright$\ }{}
\SetKwInput{KwInit}{Initialise}

\usepackage{cleveref}                                        %
\crefname{assumption}{assumption}{assumptions}
\crefname{problem}{problem}{problems}
\crefname{algorithm}{Alg.}{Algs.}
\Crefname{algorithm}{Algorithm}{Algorithms}
\crefname{figure}{Fig.}{Figs.} %
\crefformat{equation}{(#2#1#3)}
\crefrangeformat{equation}{(#3#1#4) to~(#5#2#6)}
\crefmultiformat{equation}{(#2#1#3)}%
{ and~(#2#1#3)}{, (#2#1#3)}{ and~(#2#1#3)}

\usepackage[
    backend=biber
    ,giveninits=true                  %
    ,url=false, isbn=false, doi=false %
    ,maxnames=99                       %
    ,minnames=3                       %
    ,eprint=false            %
    ,sorting=none             %
    ,date=year                %
    ,labeldate=year
    ,style=ieee
]{biblatex}
\AtEveryBibitem{                      %
    \iffieldequalstr{eprinttype}{jstor}
    {\clearfield{eprint}}
    {}
}

\AtEveryBibitem{%
    \clearfield{note}%
    \clearfield{keywords}%
    \ifentrytype{inproceedings}{%
        \clearfield{labelyear}%
        \clearfield{publisher}%
        \clearfield{volume}%
        \clearfield{urldate}%
        \clearfield{date}%
        \clearfield{pages}%
    }{%
    }%
}
\DeclareSourcemap{
    \maps{
        \map{
            \pertype{inproceedings}
            \step[fieldset=publisher, null]
            \step[fieldset=urldate, null]
            \step[fieldset=volume, null]
            \step[fieldset=pages, null]
        }
    }
}
\renewrobustcmd*{\bibinitdelim}{\,} %
\Crefname{figure}{Fig.}{Figs.}

\title{Learning to Plan Optimally with Flow-based \\ Motion Planner}

\author{
    Tin Lai\\
    School of Computer Science\\
    The University of Sydney\\
    Australia\\
    \texttt{tin.lai@sydney.edu.au} \\
    \And
    Fabio Ramos\\
    School of Computer Science\\
    The University of Sydney\\
    and NVIDIA USA\\
    \texttt{fabio.ramos@sydney.edu.au} \\
}

\begin{document}

\maketitle
\thispagestyle{empty}
\pagestyle{empty}

\begin{abstract}
    Sampling-based motion planning is the predominant paradigm in many real-world robotic applications, but 
    its performance is immensely dependent on the quality of the samples.
    The majority of traditional planners are inefficient as they use uninformative sampling distributions as opposed to exploiting  structures and patterns in the problem to guide better sampling strategies. Moreover, most current learning-based planners are susceptible to posterior collapse or mode collapse due to the sparsity and highly varying nature of \emph{C-Space} and motion plan configurations.
    In this work, we introduce a conditional normalising flow based distribution learned through previous experiences to improve sampling of these methods.
    Our distribution can be conditioned on the current problem instance to provide an informative prior for sampling configurations within promising regions.
    When we train our sampler with an expert planner, the resulting distribution is often near-optimal, and the planner can find a solution faster, with less invalid samples, and less initial cost.
    The normalising flow based distribution uses simple invertible transformations that are very computationally efficient, and our optimisation formulation explicitly avoids mode collapse in contrast to other existing learning-based planners.
    Finally, we provide a formulation and theoretical foundation to efficiently sample from the distribution; and
    demonstrate experimentally that, by using our normalising flow based distribution, a solution can be found faster, with less samples and better overall runtime performance.
    
\end{abstract}

\section{Introduction}

One of the most fundamental questions in robotics is how to manoeuvre a robot from its current position to a target position.
This problem exists in many applications, ranging from simple vacuum robot to advanced surgical manipulators.
If an autonomous robot needs to interact with its surrounding physical environment, the robot needs a method to plans for a safe trajectory that can transit itself from its current state to its desire target state.
Such an approach is called motion planning, and aims to produce a feasible trajectory that is collision-free while favouring specific paths over others by a given criterion such as distance travelled or energy consumption.

Robots operate in a different space than humans---robot configurations, or states---that are transitable is encapsulated by the term \Cspace (configuration space)~\autocite{kavraki1996_AnalProb}.
More precisely, \Cspace denotes the set of all possible configurations and is often characterised by the degrees of freedom in a given robot.
More joints will provide greater flexibility to a robot; however, this comes with increasing computational complexity. 
As the volume of the \Cspace is exponential in the number of dimensions of the robot, performing an exhaustive search becomes computationally intractable as the number of dimensions increases.
Indeed, the curse of dimensionality has been widely regarded as the main reason for prohibiting one to apply classical $A^*$ algorithms on environments or robots with many degrees of freedom~\autocite{russell2016_ArtiInte}.
Sampling-based motion planners (SBPs) represent an alternative class of methods that address these issues with random sampling strategies~\autocite{elbanhawi2014_SampRobo}.
Rather than explicitly constructing a \Cspace, SBPs build a graph or tree-like structure that captures valid trajectories between different configurations within the space~\autocite{lavalle2006_PlanAlgo}.
Its expansive nature allows it to rapidly finds a solution, and its probabilistic nature allows it to always find a solution, when one exists, given sufficient time.

SBPs are robust and applied in a substantial number of real-world applications~\autocite{macfarlane2003_JerkMani,goerzen2010_SurvMoti}. However, its original formulation uses uninformed random samplings which are often inefficient~\autocite{elbanhawi2014_SampRobo}.
Intuitively, if the \Cspace that robot operates in is purely random in nature, then using a uniform sampling scheme is justifiable.
However, in almost all real-world application, the underlying \Cspace structure, despite being complex, still retains some exploitable patterns inherent from the corresponding workspace.
For example, consider the \Cspace of a robotic manipulator illustrated in~\cref{fig:robotic-arm} for a randomly constructed workspace.
While the \Cspace seems distinct to its workspace, remaining patterns are still visible and can be exploited through learning.

\textbf{Contributions:} In this work, we introduce flow-based motion planners---PlannerFlows---that utilises a learning-based sampling distribution via normalising flows.
The learned conditional distribution allows our planner to leverage previous experience by drawing learned samples from prior optimal motion plans.
Unlike similar approaches, our flow-based distribution is robust against overfitting as its optimisation objective directly minimises mode collapse.
Since each layer of the flow is a simple bijective transformation, there is barely any runtime impact compared to similar approaches.
Furthermore, our approach is also useful in explorative tasks where there is no well-defined target state, as our conditional distribution can provide grounded inference when incomplete information are given.
Experimental results show that PlannerFlow outperforms other state-of-the-art approaches by finding a solution in less time, and often with a trajectory that is closer to the optimal.

\section{Background}

\subsection{Preliminary}
Let $\C \subseteq \mathbb{R}^d$---the \Cspace---be the set of all possible robot configurations, where $d \ge 2$ is the dimensionality of the space.
Subsequently, we will use $\Cobs \subseteq \C$ to denote the set of all invalid states (e.g. due to environment obstacles, joint limits, etc), and define the set of valid states $\Cfree$ as the closure set of $\Cfree := \text{cl}(\C\setminus\Cobs)$.
In the SBPs problem, an agent is given a starting state $\qinit$ and a target state $\qtarget$, with the objective to construct a trajectory that connects \qinit to \qtarget.
Such a trajectory $\sigma$ is a sequence of consecutively connected configurations $\sigma : [0,1]$ where $\sigma(\tau) \in \Cfree, \forall \tau \in [0,1]$.
That is, a valid trajectory that completes the agent's objective must be entirely collision free, along with $\sigma(0) = \qinit$ and $\sigma(1) = \qtarget$.

\begin{problem}[Asymptotic optimal planning]\label{problem:optimal-planning}
Given 
$C$, 
$\Cobs$, 
a pair of $\qinit$, $\qtarget$ 
and a cost function $\mathcal{L}_c : \sigma \to [0, \infty)$; the optimal motion planning problem is defined as follow.
Let $\Gamma(\Cfree)$ denote the set of all possible trajectories in $\text{cl}(\Cfree)$.
Then, an optimal trajectory $\sigma^* : [0,1] \to \Cfree$ is obtained by minimising $\mathcal{L}_c$ such that $\sigma^*(0) = \qinit$, $\sigma^*(1) = \qtarget$, and $\mathcal{L}_c(\sigma^*) = \min_{\sigma \in \Gamma(\Cfree)} \mathcal{L}_c(\sigma)$.
That is, the optimal solution is a feasible (safe) trajectory that incurs the lowest cost.
\end{problem}

A classical SBP agent constructs its trajectory by sampling possible configurations from some sampling distribution $p(q)$, and then extends its closest existing node from its tree towards the sampled configuration.
For ease of expression, we will normalise each dimension of $C$ into a unit length of one, such that the \Cspace is given by $[0,1]^d$.
Then, a SBP agent typically samples random configurations $q$ from a uniform distribution $q \sim \mathcal{U}(0, 1)^d$ for its tree extensions.
This process is repeated to find a valid trajectory that satisfies \cref{problem:optimal-planning}.
The trajectory $\sigma^*$ will be continuously refined to the optimal trajectory.

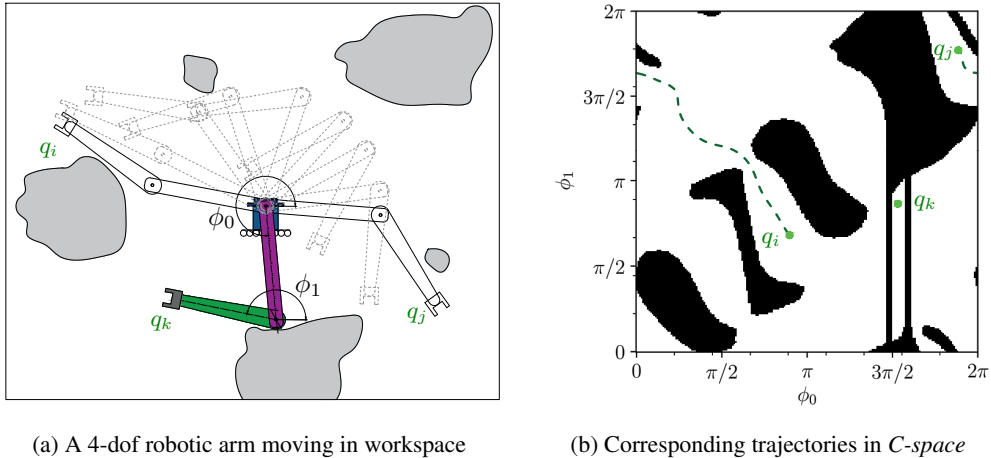
\begin{figure}[tb]
    \centering
    \subcaptionbox{%
        A 4-dof robotic arm moving in workspace \label{fig:robotic-arm:obstacle-space}}%
    {%
        \begin{tikzpicture}
            \node (img) {\includegraphics[height=0.24\textheight]{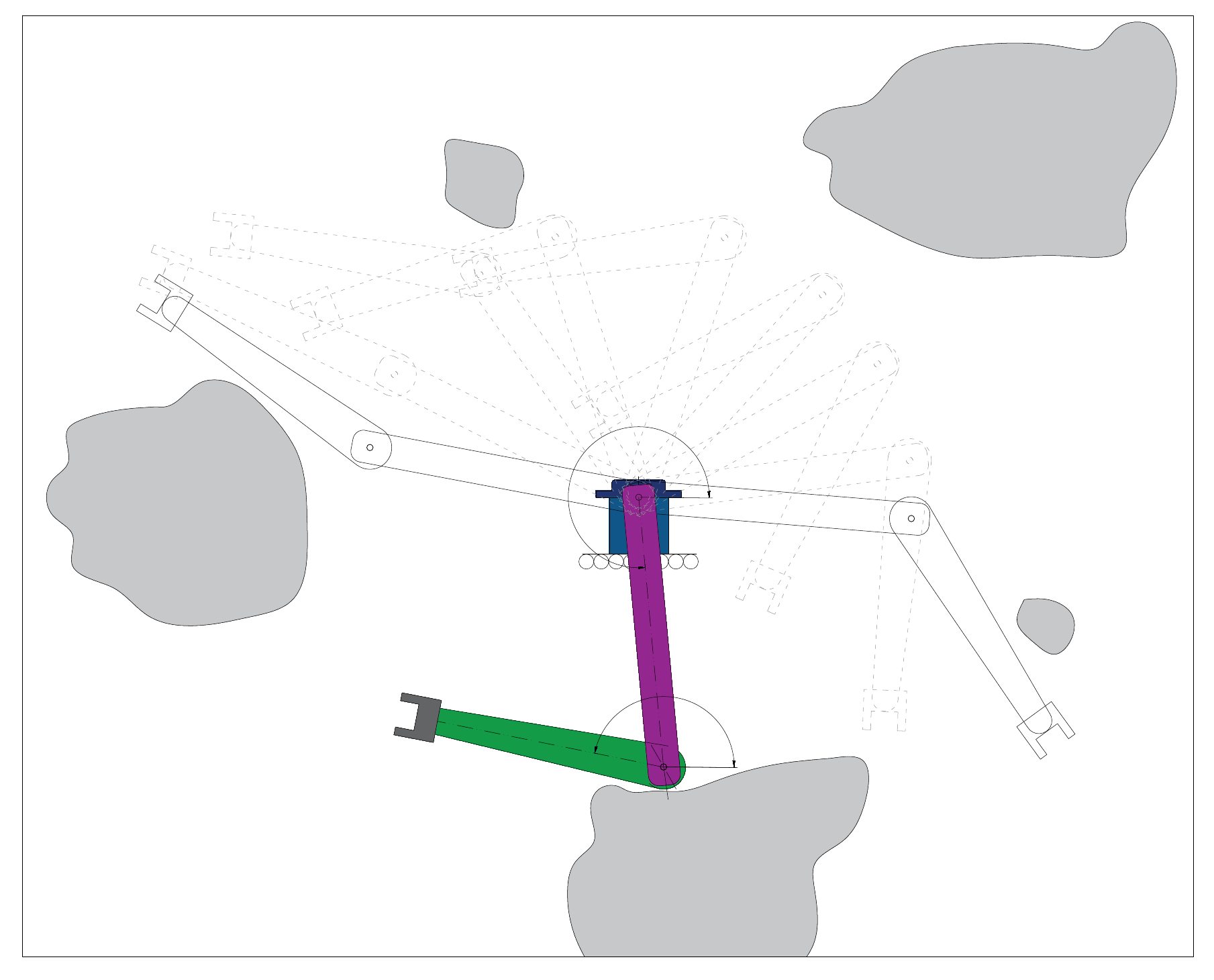}};

            \node[xshift=-.4cm, yshift=-.2cm] (t1) {$\phi_0$};
            \node[align=left,xshift=.75cm, yshift=-1.1cm] (t2) {$\phi_1$};

            \node at (-2.7, .7) (q1)    {\color{green!50!black}\small$q_i$};
            \node at (2.2, -1.5) (q2)    {\color{green!50!black}\small$q_j$};
            \node at (-1.2, -1.6) (q3) {\color{green!50!black}\small$q_k$};
        \end{tikzpicture}
    }
    \hspace{.1cm}
    \subcaptionbox{%
        Corresponding trajectories in \Cspace
        \label{fig:robotic-arm:configuration-space}}%
    {%
        \begin{tikzpicture}
            \node (img) {\includegraphics[height=0.24\textheight]{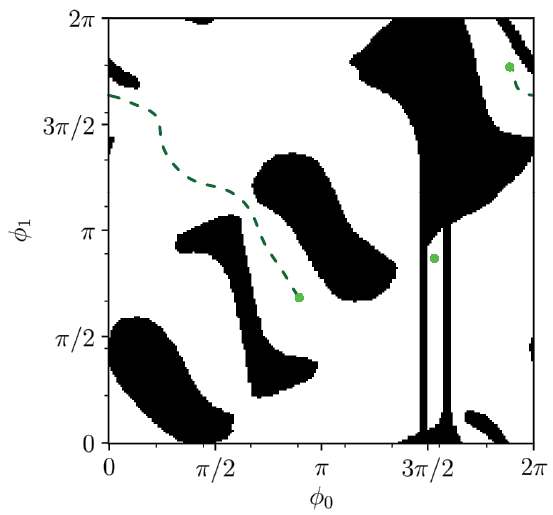}};
            
            \node at (0, -.5) (q2)    {\color{green!50!black}\small$q_i$};
            \node at (2.3, 2) (q1)    {\color{green!50!black}\small$q_j$};
            \node at (2.05, 0) (q3) {\color{green!50!black}\small$q_k$};
        \end{tikzpicture}%
    }
    \caption{
         Illustration of planning a trajectory from $q_i$ to $q_j$ for a robotic arm  with 4-degree of freedom (dof)---2 rotational joints $\{\phi_i\}^1_{i=0}$ and 2 dimensional movement, in (\subref{fig:robotic-arm:obstacle-space}) workspace and (\subref{fig:robotic-arm:configuration-space}) \Cspace.
        While there is significantly more free space in the workspace view, the \Cspace view demonstrates that planning over feasible configurations is very cluttered and therefore more challenging.
        There exists numerous narrow passages and inaccessible areas.
        Note that the movement is fixed for the 2D projection, and the \Cspace view is shown solely for illustration purposes and is not available during planning.
        \label{fig:robotic-arm}
    }
\end{figure}

\subsection{Related works}

Although the \emph{probabilistically completeness} ensures that the use of uniform sampling will eventually leads to a solution~\autocite{elbanhawi2014_SampRobo}, it is uninformative and, in most cases, extremely inefficient.
The inefficiency is even more prominent in spaces with narrow passages~\autocite{hsu1998_FindNarr} (e.g. there are several in~\cref{fig:robotic-arm:configuration-space}), which substantially limits the connectivity of \Cfree.
That is because extending a connection within the narrow passages correlates to the area of free spaces within the narrow passages~\autocite{lai2018_BalaGlob}; a uniform sampling distribution will have very low probability due to limited free spaces.
\Cref{fig:robotic-arm} illustrates the trajectories of a manipulator in a workspace translated to configuration space, where there exist exploitable structures.
It is impossible to explicitly plan for the underlying \Cspace; however, there exist different approaches which implicitly tackle this issue, including techniques such as adaptive sampling~\autocite{yershova2005_DynaRRTs}, heuristic measures of obstacles boundary~\autocite{zhang2008_EffiRetr,lee2012_SRRRSele}, and learning-based approaches to modify the sampling distribution.

There are several prior works that aim to tackle the sampling efficiency problem in SBPs~\autocite{sun2005_NarrPass,gammell2018_InfoSamp,lai2019_LocaSamp}. Traditionally, most of them are rule-based methods designed to address the narrow passage issue~\autocite{hsu1998_FindNarr}.
Rule-based methods do not scale well with \Cspace's increases in complexity~\autocite{urmson2003_ApprHeur} and often neglect any prior knowledge as each run is a fresh start.
In contrast, learning-based methods consider their previous experience in similar environments and are often able to perform better than rule-based or uninformed planners~\autocite{ichter2018_LearSamp}.
There are attempts to directly use learning techniques to construct motion plans, such as potential fields~\autocite{hwang1992_PoteFiel,koren1991_PoteFiel}, reinforcement learning~\autocite{sartoretti2019_PRIMPath}, and encoder-decoder networks~\autocite{qureshi2019_MotiPlan}.
These approaches directly use learning-based methods to replace the constructions of motion plans entirely from prior experience; however, such approaches are often limited in scope or require some fallback planner to handle failed cases~\autocite{qureshi2019_MotiPlan}.

The quality of sampling distribution is critical in SBPs as the growth of its roadmap depends on it.
There is an extensive literature on estimating complex distributions,
but they are not always applicable to motion planning due to issues like runtime performance~\autocite{everett2018_MotiPlan}, loss of completeness guarantee~\autocite{rickert2008_BalaExpl}, or falling into the same common pitfall of mode collapse in generative models~\autocite{qureshi2019_MotiPlan}.
Modelling sampling distribution can improve the efficiency of SBPs without compromising \emph{completeness}~\autocite{lai2020_BayeLoca}, yet it remains an open question how to find a suitable learning-based procedure to effectively utilise prior experience.
Encoder networks are a class of promising models that can significantly enhance the sampling efficiency in SBPs~\autocite{ichter2018_LearSamp}, yet standard techniques like variational autoencoders~\autocite{kramer1991_NonlPrin} are prone to mode collapse due to the target distribution \Cspace's sparsity.
In our flow-based motion planner, we stack a series of simple, invertible transformation layers known as normalising flow~\autocite{tabak2013_FamiNonp} that transforms a simple Gaussian into a complex distribution. The transformation network learns the mapping between the environment and optimal motion plans.
We then compose coupling blocks to allow our flow-based distribution to be conditioned on planning information, incorporating contextual clues.
The final conditional distribution is easy to compute, and can be trained by optimising a loss that minimises mode collapse.

\section{Motion planning with normalising flow sampling}

\subsection{Parameterising distribution for motion planning}

Sampling-based motion planner extends its roadmap tree via sampling random configurations from some distribution $q_\text{rand} \sim p(q)$---most commonly a uniform distribution $\mathcal{U}_q(0,1)^d$ extended along the \Cspace.
While sampling from a uniform distribution is probabilistically complete, it ignores vital spatial information which is crucial to motion planning.
In this work, we construct a normalising flow-based distribution $\SampleDist$ that maximises the likelihood of sampling configurations close to the optimal trajectory.
This is achieved by conditioning $\SampleDist$ on information specific to the motion planning problem.
$\SampleDist$ is a parametric function of $p$ that implicitly learns the encoding between a planning problem and its optimal trajectory region.

The most common information available in any given planning problem includes the workspace information, the starting configuration and the target configuration.
Let $\workspace$ denote a finite dimensional encoding of the workspace information. We can rewrite the conditional distribution as $\SampleDist(q \given \workspace, \qinit, \qtarget)$.
For learning such a conditional distribution from empirical data, we will formulate the distribution as a latent variable model in a parameterised form $p_\theta(q \given z, \workspace, \qinit, \qtarget)$, where $z$ is a latent variable, and $\theta$ is a vector of parameters.
We will then maximise the likelihood,
\begin{equation}\label{eq:distribution-Q}
    \SampleDist_\theta(q \given \workspace, \qinit, \qtarget) = \int p_\theta(q \given z, \workspace, \qinit, \qtarget)\, p_\theta(z \given \workspace, \qinit, \qtarget) \diff z,
\end{equation}
with respect to the empirical distribution.
We will discuss modelling the joint distribution with a conditional invertible normalising flow in the following sections.
The empirical distribution is to be taken from expert demonstrations, which could either be human demonstration (for mimicking human behaviour) or from some optimal motion planner that maximises its objective function.

Drawing samples for motion planning requires a computationally efficient distribution because sampling is the basis of SBPs---a moderate problem often already requires several hundred thousands of sampled configurations to complete.
Since we want to utilise information from the robot's surroundings, the distribution needs to be able to condition on the planning information.
In addition, it should be robust against mode collapse, which is very common in generative models as an SBP with a collapsed distribution would suffer severely in its planning capability.
These are common problems associated with existing related techniques, and our proposing framework aims to tackle these issues by learning a complex conditional distribution with normalising flows.

\subsection{Modelling the sampling distribution with normalising flows}

Normalising flows is a class of methods typically used to approximate a posterior distribution in variational inference~\autocite{rezende2015_VariInfe}.
It describes the transformation of a probability density through a sequence of invertible mappings, where the input density, typically Gaussian or uniform, \emph{flows} through the mappings and is transformed into another distribution.
The method is quite flexible in its representational capacity and can be used to approximate arbitrarily complex posterior distributions~\autocite{lu2019_StruOutp}.
It can also be extended to conditional distributions resulting in models similar to a conditional GAN~\autocite{mirza2014_CondGene} and CVAE~\autocite{sohn2015_LearStru}. 
However, a normalising flow learns the data distribution $p(x)$ explicitly, which is more suitable for motion planning as it is less prone to mode collapse.

Let $z_0$ be some random variable with a distribution $p(z_0)$ that is analytically tractable, for example a Gaussian.
A normalising flow $f=(f_1,\ldots,f_k,\ldots,f_K)$ is a mapping from this initial density $z_0$ to a target density $z_K$ through a sequence of transformations
\begin{equation}\label{eq:zk-to-z0}
    z_K = f_K(f_{K-1}(\ldots f_1(z_0) )),
\end{equation}
where $K$ is the number of layers within the flow, and $f_k$ denotes a single transformation layer.
Here and thereafter we will use the notation $z_K = f_K \circ f_{K-1} \circ \cdots \circ f_1(z_0)$ to representation the corresponding composition in~\cref{eq:zk-to-z0} for brevity.
Each transformation layer consists of a smooth mapping $f_k : \mathbb{R}^d \mapsto \mathbb{R}^d$ that is invertible where $g_k = f_k^{-1}$ exists and $g_k\circ f_k(z) = z$.
If we map $z$ with distribution $p(z)$ to $f_k$, then the resulting random variable $z'=f_k(z)$ has distribution
\begin{equation}\label{eq:layer-transform}
    p(z') = p(z) 
        \left|
        \mathrm{det} \frac{\partial f_k^{-1}}{\partial z'}
        \right|
        =
        p(z)
        \left|
        \mathrm{det} \frac{\partial f_k}{\partial z}
        \right|^{-1}
\end{equation}
by applying the chain rule and Jacobians property of invertible functions.
Therefore, we can construct arbitrarily complex densities by successively applying~\cref{eq:layer-transform} as a mapping to its previous layer's output.
This forms the basis of the \emph{flow}.
By applying a chain of $K$ transformations we can rewrite~\cref{eq:zk-to-z0} as
\begin{equation}
    \ln p(z_K) = \ln p(z_0) - \sum^K_{k=1} \ln
    \left|
        \mathrm{det}
        \left(
            \frac{\diff f_k(z_{k-1})}{\diff z_{k-1}}
        \right)
    \right|,
\end{equation}
which is the log density of $z_K$.

We use a standard Gaussian distribution for the initial distribution $p(z_0)$, and each $f_k \in \set{f_k}_{k=1}^{K}$ is a simple, monotonically increasing function with closed-form derivative.
Each layer transforms the input probability density with the operation that expands or contracts the density distribution.
These sequence of operations enable us to model more complex representations of the target distribution. This is in contrast to symmetric distributions such as the initial Gaussian that fail to capture multi-modality usually appearing in \Cspace.
Normalising flow allows us to use a transformation $f_k$, which is individually simple and easy to compute~\autocite{rezende2015_VariInfe} yet can be arbitrarily flexible by stacking more layers.

\subsection{Conditioning on motion plan information}\label{sec:conditioning-on-motion-plan-info}

The conditional distribution in our PlannerFlows framework uses the affine coupling block architecture for coupling the planning information~\autocite{ardizzone2019_GuidImag}.
Each transformation function $f_k$ learns a bijective model which is both tractable and extremely flexible.
Together with our flow $f$, it stacks a sequence of bijection blocks capturing the nonlinearity and complexity of the density.
Our network is conditioned on the planning information $\workspace$, $\qinit$, and $\qtarget$---which constitute an abstract representation of the problem instance.
Each coupling block with  input $z_k$ is split into two components $[z_{k,a}, z_{k,b}]$, where
\begin{alignat}{3}
    z_{k+1,a} &= z_{k,a} \odot \exp (s_a (z_{k,b} \given \workspace, \qinit, \qtarget )) &+& t_a (z_{k,b}\given\workspace, \qinit, \qtarget) \\
    z_{k+1,b} &= z_{k,b} \odot \exp (s_b (z_{k+1,a}\given\workspace, \qinit, \qtarget)) &+& t_b (z_{k+1,a}\given\workspace, \qinit, \qtarget)
\end{alignat}
applies affine transformations between the split components, which have strictly upper or lower triangular Jacobians.
Each $s$ and $t$ are scale and translational functions that map from $\mathbb{R}^d \mapsto \mathbb{R}^{D-d}$ where $D$ is the original dimensionality of $z_k$, and $d$ is the dimensionality of $z_{k,a}$.
They can be represented by arbitrary neural networks and conditioned on our planning information $\workspace, \qinit, \qtarget$ when performing the coupling.
Note that since $z_{k,a}$ and $z_{k,a}$ are triangular Jacobians, the determinant of these triangular matrices can be computed efficiently as the product of its diagonal terms.
The splitting and subsequent coupling enable us to efficiently compute each layer $f_k$, making each transformation layer inexpensive to compute.
The same process is repeated throughout the network $f$, and the output of each coupling block will then be concatenated and passed to the next block.

\subsection{Training motion sampler with prior experience}

Our flow-based distribution is given by $f_\theta$--- parameterised by the parameter $\theta$ and conditioned on the planning information.
We can rewrite~\cref{eq:distribution-Q} in terms of our model as 
\begin{equation}\label{eq:sampdist-Q-given-planninginfo}
    \SampleDist_\theta(q \given \workspace, \qinit, \qtarget) = p_Z(f_\theta(q \given \workspace, \qinit, \qtarget))
    \left|
        \mathrm{det} \frac{\partial f}{\partial q}
    \right|,
\end{equation}
where $Z$ is the latent space of our latent variable z, and ${\partial f}/{\partial q}$ is the Jacobian matrix.
We can then substitute the posterior over model parameters $p(\theta \given q, \workspace, \qinit, \qtarget) \propto \SampleDist_\theta(q \given \workspace, \qinit, \qtarget)p(\theta)$ from Bayes' theorem into~\cref{eq:sampdist-Q-given-planninginfo}.
This is the classical {\em maximum a posteriori} fitting of parameters $\theta$ that minimises the loss
\begin{equation}\label{eq:loss-distribution}
    \mathcal{L} = \mathbb{E}_i 
                [ -\log(\SampleDist_\theta(q_i \given \workspace, \qinit, \qtarget)) ]
                - \log (p(\theta))
\end{equation}
for each training sample $q_i$.
In this work, we use the Gaussian distribution as our base distribution $p_Z(z)$, with Gaussian prior on weights $\theta$.
By rewriting~\cref{eq:loss-distribution} and substituting our model $f_\theta$, we obtain
\begin{equation}\label{eq:loss-rewrite}
    \mathcal{L} = \mathbb{E}_i
                \left[
                \frac{
                    \left\lVert
                        f_\theta(q_i \given \workspace, \qinit, \qtarget)
                    \right\rVert^2_2
                }{2}
                - \log \big| \mathbf{J}_i \big| 
                \right]
                + 
                \frac{
                    \lVert \theta \rVert^2_2
                }{
                    2\sigma^2_\theta
                },
\end{equation}
where $\mathbf{J}_i := \mathrm{det}(\partial f / \partial q |_{q_i})$ denotes the determinant of the Jacobian.
The former term in~\cref{eq:loss-rewrite} represents the maximum likelihood loss while the latter is a regularisation penalty.

Our flow-based distribution is to be used in place of an uninformative distribution.
The distribution of the training data will then define the behaviour of $\SampleDist_\theta$; for example, training with a motion planner that has certain routing preference will causes the learned distribution to inherent such behaviour.
For any given robot, its range of free movements are often fixed, but it has to execute manipulation in varying environments, under different starting conditions.
Intuitively, similar environments often implies information from prior motion plans are useful in this given instance, where the amount of correlation is dependent on the deviation of conditions.
We train $\SampleDist_\theta$ with data that consists of motion plans across different workspaces.
In a given workspace, there exists several demonstrations of trajectories with varying initial and target configurations.
The demonstrations are collected with expert sampling-based motion planner that optimises Euclidean cost, and each trajectory is sparsely sampled such that only configurations that contribute to the final near-optimal trajectory are taken as the training set.
Moreover, since there will be multiple similar demonstrations in any given problem instance, it ensures $\SampleDist_\theta$ has a diverse view of the distribution.

It should be noted that while learning from such an optimal planner will allow $\SampleDist_\theta$ to inherent its planning preference, it does not automatically grants $\SampleDist_\theta$ the ability to always sample optimally.
Rather, the distribution $\SampleDist_\theta$ is not to replace the robustness of traditional SBPs, but to introduce a prior belief on the sampling space from previous experience in similar-looking environments (and similar starting conditions).
The prior belief allows SBP to learn from its experience, and focus more of its ``attention'' on more promising sampling regions.
We will demonstrate in~\cref{sec:Q-cond-incomplete-info} that $\SampleDist_\theta$ can learn from its prior experience and sample efficiently even with incomplete information.

Furthermore, training a network with loss $\mathcal{L}$ in~\cref{eq:loss-rewrite} yields an estimate of $\theta$ that is less likely to mode collapse~\autocite{winkler2019_LearLike}.
The reason is that during training, any mode in the training set with low probability under the current estimate of $\SampleDist_\theta(q_i \given \workspace, \qinit, \qtarget)$ will have a latent vector that lies far outside the normal distribution $p_Z$, which incurs a large loss for the first L2 term in~\cref{eq:loss-rewrite}.
In contrast, similar generative approaches such as GAN only provide a weak signal in every training epoch, which is inclined to discard low occurring modes completely.
Such a property is essential for motion planning, as the distribution of trajectory configurations is inherently sparse, with irregular structure.

\vspace{-1mm}
\section{Experiments}
\vspace{-2mm}

{
    \setlength\intextsep{0pt}
    \captionsetup{belowskip=0pt}
    \setlength{\belowcaptionskip}{-10pt}
    \begin{wrapfigure}{r}{.34\linewidth} %
        \centering
            \includegraphics[width=\linewidth,frame]
            {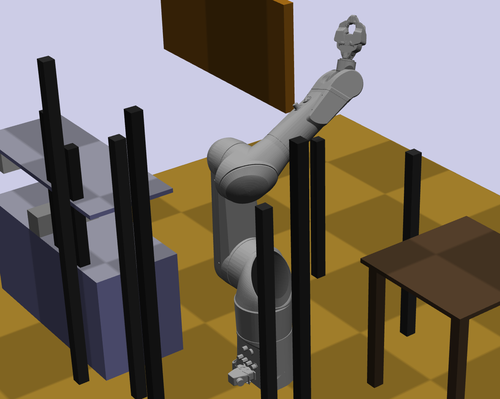}
        \caption{\footnotesize
            Example of the 6D manipulator environment with the TX90 robot arm with an attached PR2 gripper.
            \label{fig:tx90}
        }
    \end{wrapfigure}

    Three main types of environments were tested---a 2 dof point mass robot, a 4 dof movable robot arm in 2D workspace, and a 6 dof robotic manipulator in 3D workspace.
    In order to learn the encoding in our conditional distribution $\SampleDist_\theta$, numerous demonstrations were performed for a wide range of similar random environments.
    To this end, environments were procedurally-generated, and subsequently, solved for random pairs of optimal motion planning problems within the same environment.
    For the 2D workspace environments (example in~\cref{fig:samp-conditioning}), we generate environments each with a random ratio $r_\text{obs}$ between the $\Cobs$ and $\Cfree$. %
    For the 3D workspace (example in~\cref{fig:tx90}), environments were generated in a more structured way (e.g. random number of poles and tables) with physical constrains (e.g. objects are enforced to be in contact with a surface).

For each of the scenarios described, 100 maps were randomly generated with each map consisting of around 200 pairs of \qinit, \qtarget motion samples.
Each scenario was trained for $1500$ epochs until convergence.
During testing, our learned flow-based distribution $\SampleDist_\theta$ was used for sampling configurations by conditioning on the current unseen environment information.
An obstacle point cloud representation of the workspace is used for $\workspace$.
We follow an approach similar to epsilon greedy, where $\SampleDist_\theta$ is mixed with an unbiased uniform distribution with $\epsilon=0.1$ as the probability of sampling from the unbiased distribution.
This approach enables our modified motion planner inherent the theoretical \emph{probabilistic completeness} guarantee from the classical SBPs, while empirically attaining a much more efficient runtime.}

\begin{figure}
\begin{floatrow}
\ffigbox[.76\textwidth]{%
        \begin{subfigure}[t]{.353\linewidth}
            \centering
            \includegraphics[width=\linewidth]
            {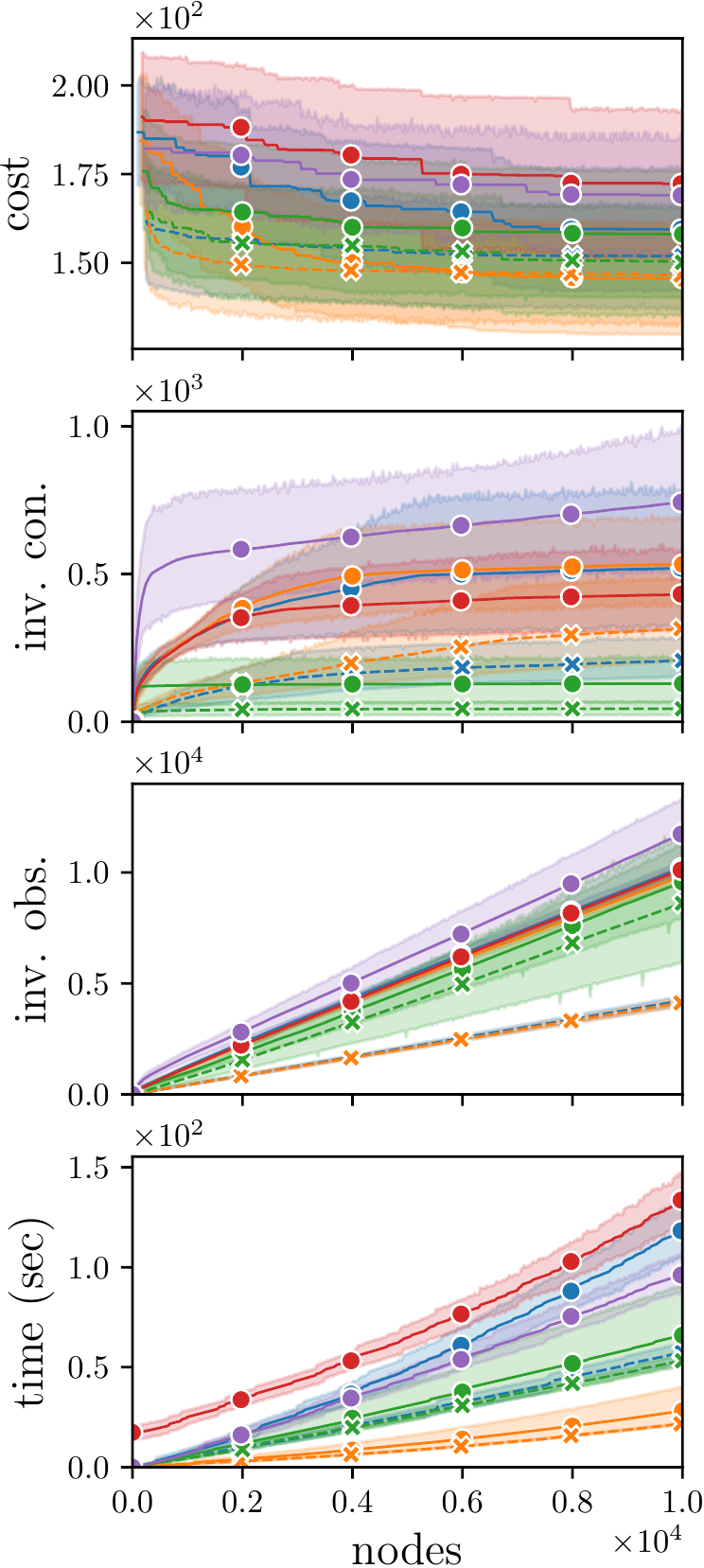}
        \caption{2D point robot\label{fig:experiments:2d}}
        \end{subfigure}%
        \begin{subfigure}[t]{.32\linewidth}
            \centering
            \includegraphics[width=\linewidth]
            {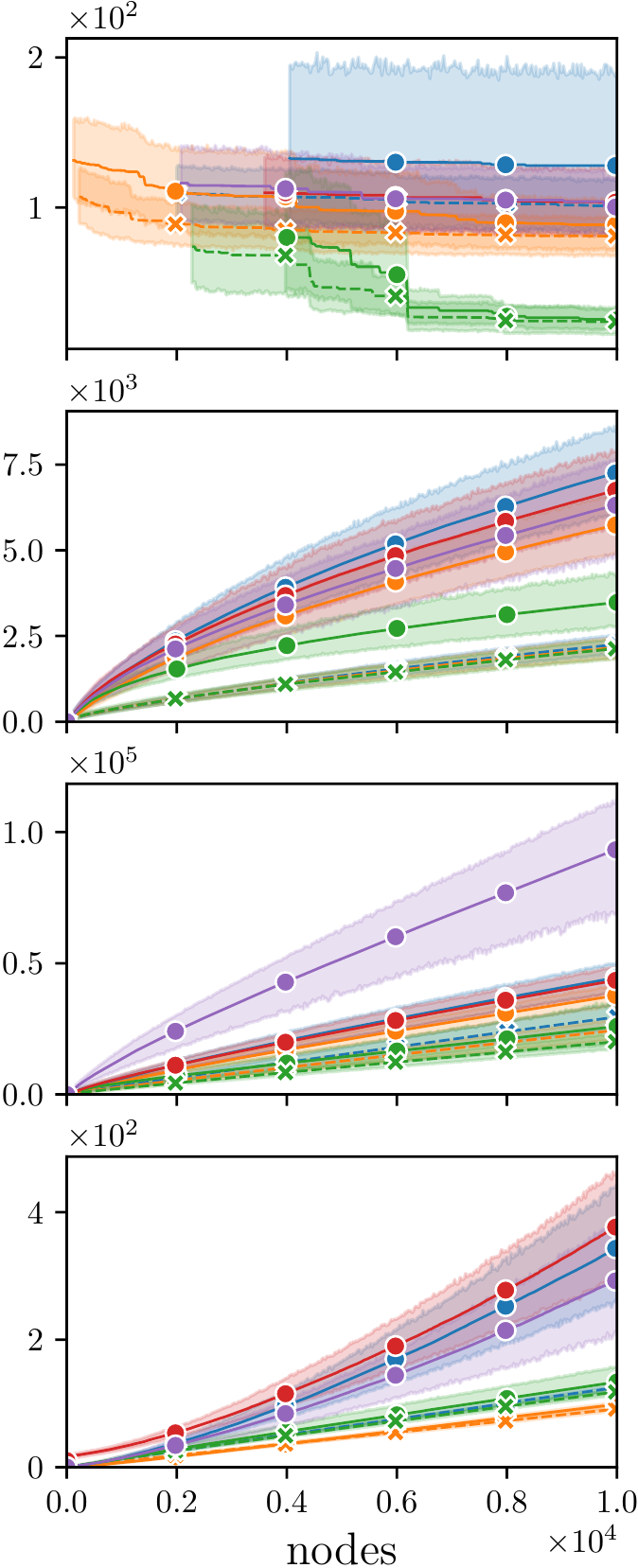}
        \caption{4D movable manipulator\label{fig:experiments:4d}}
        \end{subfigure}%
        \begin{subfigure}[t]{.32\linewidth}
            \centering
            \includegraphics[width=\linewidth]
            {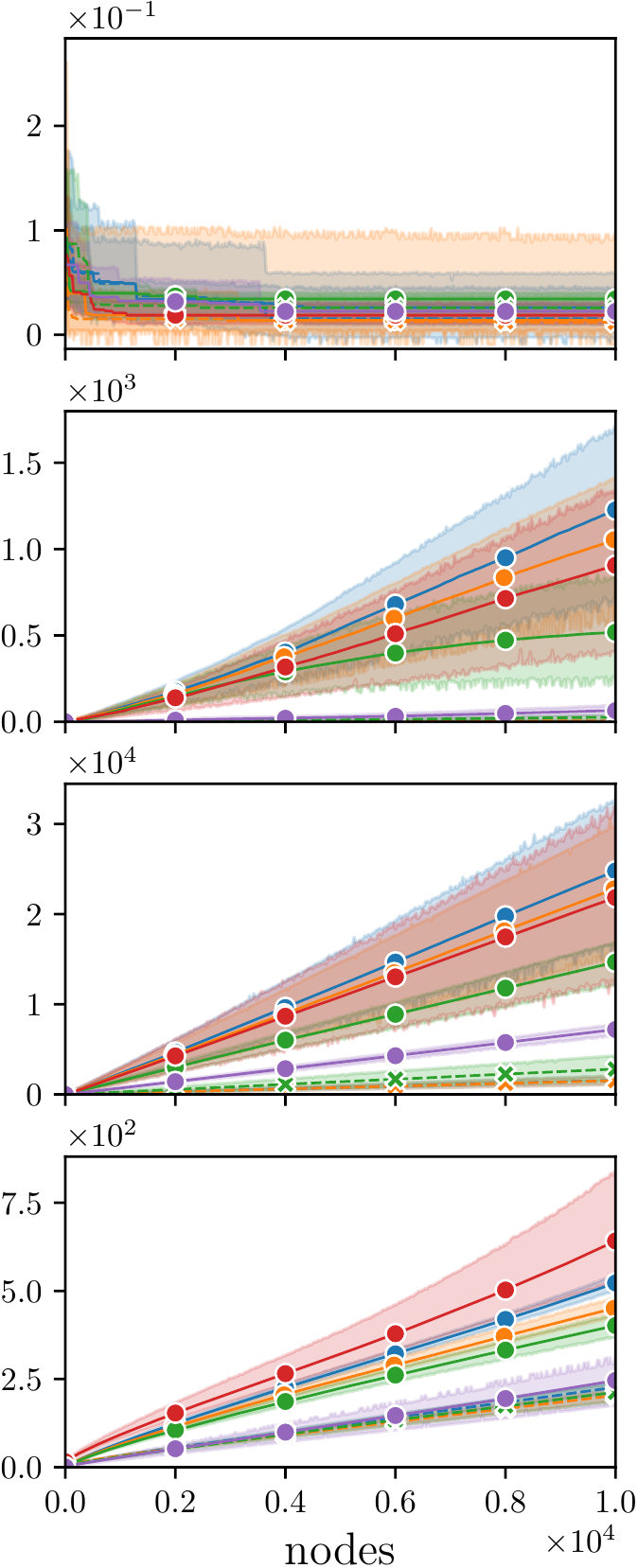}
        \caption{6D manipulator\label{fig:experiments:6d}}
        \end{subfigure}%
}{%
    \caption{\footnotesize
        Top to bottom are \emph{cost}, \emph{invalid connections}, \emph{invalid obstacles}, and \emph{time} against number of nodes.
        Each line and shaded region represents the mean and $95\%$ confidence interval respectively.
        Solid and dotted line represents the original methods, and the corresponding variant that uses our flow-based conditional distribution $\SampleDist_\theta$, respectively.
        Most dotted lines are overlapped at the bottom in~(\subref{fig:experiments:6d}).
        \label{fig:experiments}
    }
}%
\capbtabbox{%
        \begin{minipage}{.23\textwidth}
        
        \includegraphics[width=.95\linewidth,clip]
        {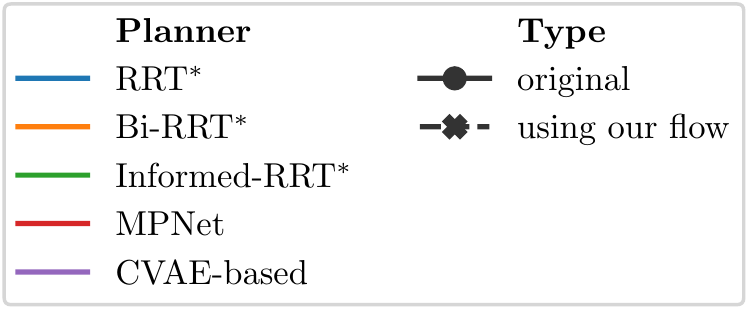}
        \vspace{\fill}
        \vspace{5mm}
        
            \tabcolsep=0.11cm
            \resizebox{\linewidth}{!}{%
                \input{expr-table}
            }

        \end{minipage}
}{%
    \caption{\footnotesize
        Total sampled configurations ($\mu\pm\sigma$) when planner is terminated at $10\,000$ nodes.
        \label{table:samp-pts}
    }
}
\end{floatrow}
\end{figure}

\vspace{-1mm}
\subsection{Results}
\vspace{-1mm}

Quantitative results are shown in~\cref{fig:experiments}.
There are 8 planners in total, including traditional SBPs---%
\RRT*~\autocite{karaman2010_IncrSamp},
bidirectional \RRT*~\autocite{kuffner2000_RRTcEffi} and 
informed \RRT*~\autocite{gammell2018_InfoSamp};
our three corresponding PlannerFlows variants that use $\SampleDist_\theta$ instead of the original sampling distribution;
and two other state-of-the-art learning-based planners---%
MPNet~\autocite{qureshi2019_MotiPlan}
and CVAE-based \RRT*~\autocite{ichter2018_LearSamp}.
The results for 2D, 4D and 6D scenarios were collected in 10 randomly picked environments, within each environment 3 different pairs of $\qinit$ and $\qtarget$, and each pair of coordinates were repeated 3 times.
The same set of planning problems were used on all planners with a budget of $10\,000$ nodes.
Experiments were performed on a machine with CPU i7-7700HQ 3.8GHz and GPU NVIDIA GTX 1050 Mobile (with 4GB of RAM).
For the \emph{cost} metric in~\cref{fig:experiments} the line begins at the first instance of not infinite cost (i.e. first time finding a solution).
The \emph{invalid obstacles} metric refers to a sampled configuration $q_\text{rand}$ being discarded because $q_\text{rand}\in\Cobs$, whereas \emph{invalid connections} refers to $q_\text{rand}\in\Cfree$ but fails to connect an existing roadmap due to intermediate collision.
\Cref{table:samp-pts} depicts the total number of sampled configurations for each planner. It represents the total number of calls on the simulator to check for collision (i.e. the lesser the better).

\vspace{-2mm}
\subsection{Performance comparison}
\vspace{-1.5mm}

Drawing samples from $\SampleDist_\theta$ improves most metrics used by SBP.
Our PlannerFlows consistently outperforms the original methods by sampling configurations that can actually be extended to its roadmap.
This is confirmed by the reduced number of invalid samples over time compared to other planners.
Moreover, since $\SampleDist_\theta$ is trained by an expert planner that produces optimal motion plans, the initial \emph{cost} tends to starts with a much lower cost than their counterpart.
Other planners need to use more nodes to reduce the cost of their trajectory, whereas $\SampleDist_\theta$ can sample in regions that are more promising to produce optimal trajectories.
Interesting, CVAE-based \RRT* performs poorly in~\cref{fig:experiments:2d,fig:experiments:4d} but outperforms traditional SBPs in~\cref{fig:experiments:6d}.
This is likely due to the highly random and sparse nature of the 2D workspace that causes the learned distribution to have collapsed modes, resulting in an drop of its sample quality.
In contrast, in the 3D workspace, obstacles are displaced in a more organised manner making the latent space comparatively easier to learn.
Note that MPNet first uses a neural network to produce a valid trajectory.
When such an attempt fails multiple times, it will then fall back to using \RRT* to fix the invalid segments. In certain scenarios, its behaviour is similar to \RRT* when its `neural planning' happens to fail in that environment.

Although PlannerFlow requires extra computation for drawing samples from the learned network, we stress the negligible performance overhead for the flow-based distribution sampling.
The normalising flow inference is a simple forward-pass in a network with monotonic functions. Results shown in~\cref{fig:experiments} were computed in a laptop-grade GPU.
We draw configuration samples in a batch size of 10\,000 (and redraws if the batch is exhausted
and the motion planning process has yet to finish), where each forward-pass only takes $0.433\pm0.011$, $0.735\pm0.010$ and $1.782\pm0.007$ second ($\mu\pm\sigma$) for the 2D, 4D and 6D environment respectively. Note that this amount of time accounts for less than 1\% of the overall runtime.
Moreover, almost all of our PlannerFlow variants take the least amount of time to finish planning, which is likely due to the high quality of samples from $\SampleDist_\theta$, saving time by reducing the number of calls to the environment simulator. This is made apparent by comparing the \emph{invalid obstacles} and \emph{invalid connections} metrics in~\cref{fig:experiments}, and the total number of sampled configurations in~\cref{table:samp-pts} where planners that use $\SampleDist_\theta$ always require less samples.
This effect is even more apparent in~\cref{fig:experiments:6d} because collision check calls with forward kinematics are more expensive in higher dimensions.

\begin{figure}[tb]
    \centering    
        \begin{subfigure}[t]{.20\linewidth}
            \includegraphics[width=\linewidth]
            {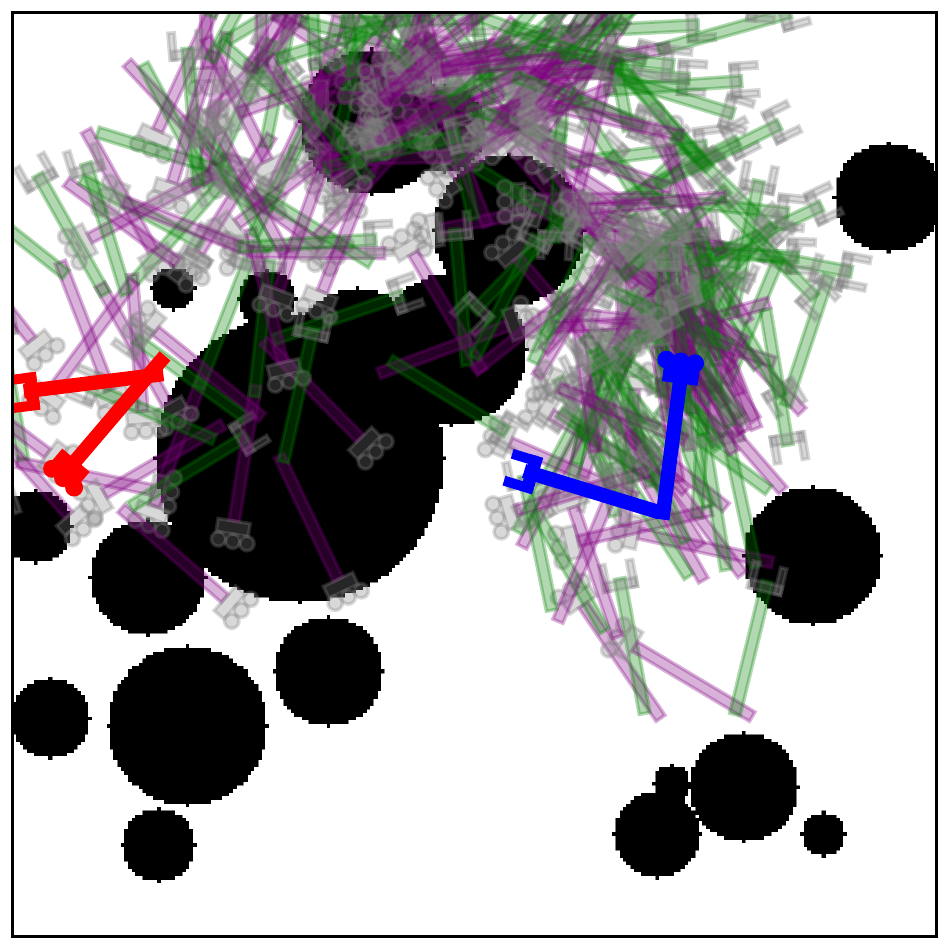}
            \caption{$\SampleDist_\theta(q \given \workspace, \qinit, \qtarget)$
            \label{fig:samp-conditioning:init-goal}
            }
        \end{subfigure}%
        \begin{subfigure}[t]{.20\linewidth}
            \includegraphics[width=\linewidth]
            {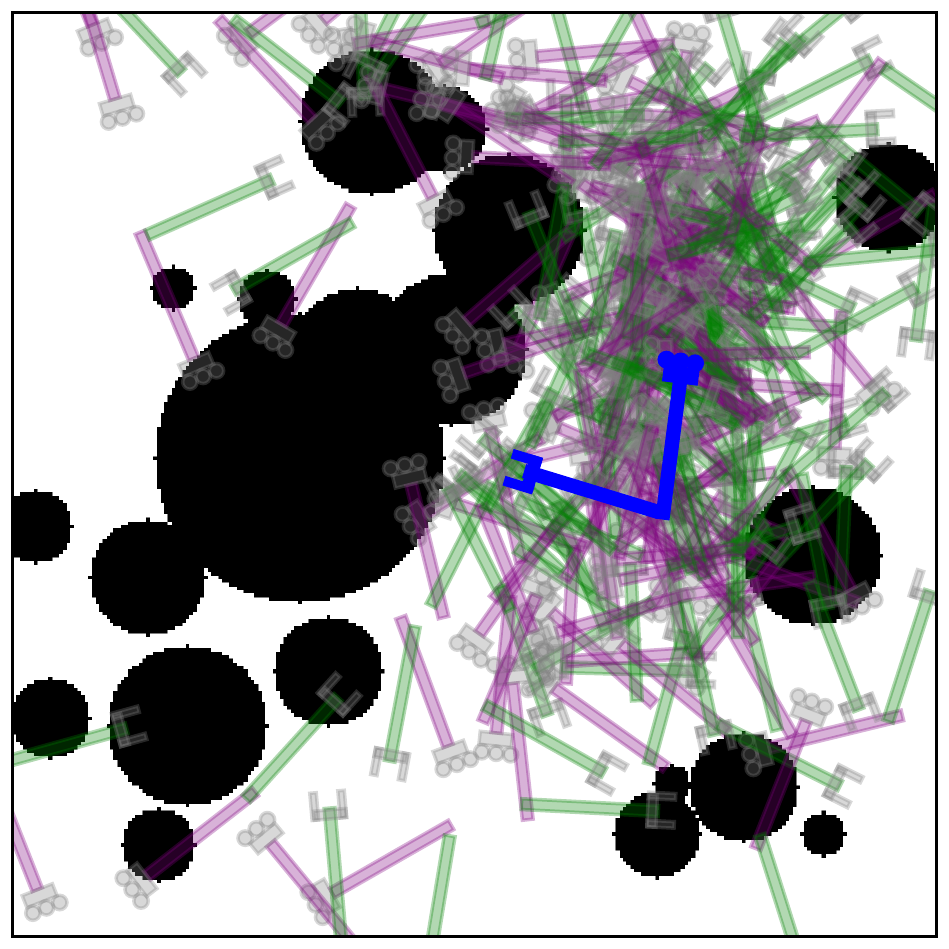}
            \caption{$\SampleDist_\theta(q \given \workspace, \qinit)$
            \label{fig:samp-conditioning:init}
            }
        \end{subfigure}%
        \begin{subfigure}[t]{.20\linewidth}
            \includegraphics[width=\linewidth]
            {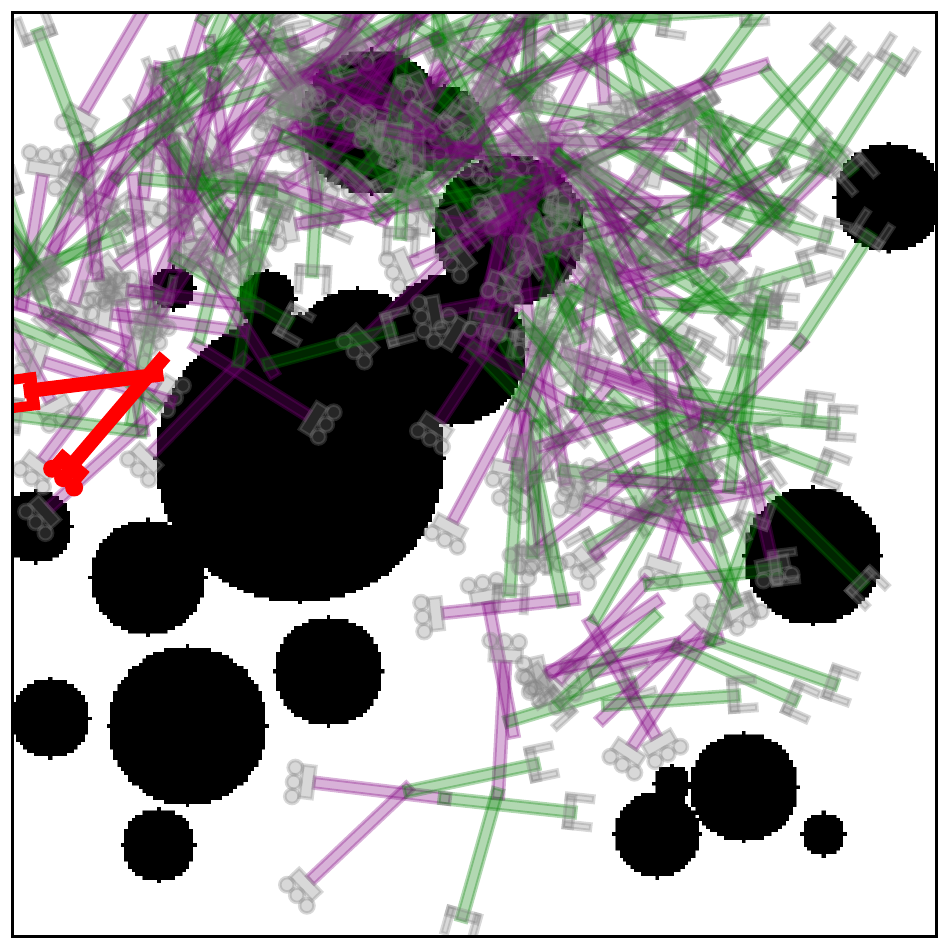}
            \caption{$\SampleDist_\theta(q \given \workspace, \qtarget)$
            \label{fig:samp-conditioning:goal}
            }
        \end{subfigure}%
        \begin{subfigure}[t]{.20\linewidth}
            \includegraphics[width=\linewidth]
            {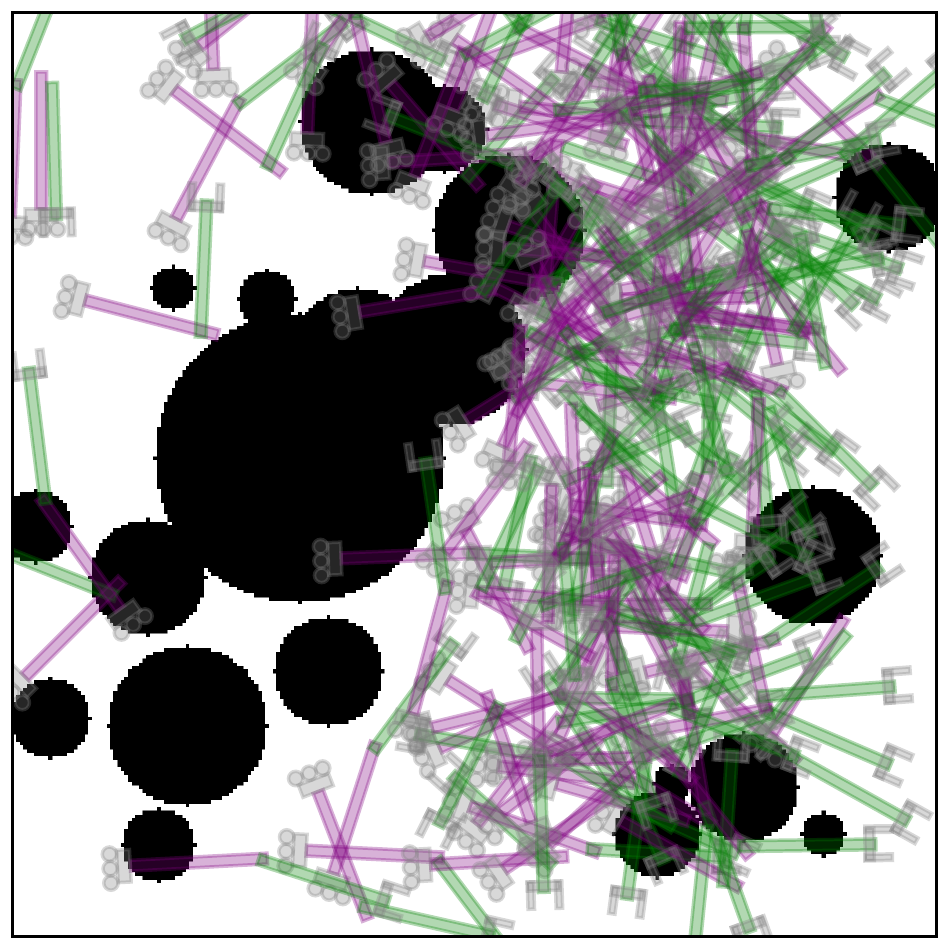}
            \caption{$\SampleDist_\theta(q \given \workspace)$
            \label{fig:samp-conditioning:none}
            }
        \end{subfigure}%
        \begin{subfigure}[t]{.20\linewidth}
            \includegraphics[width=\linewidth]
            {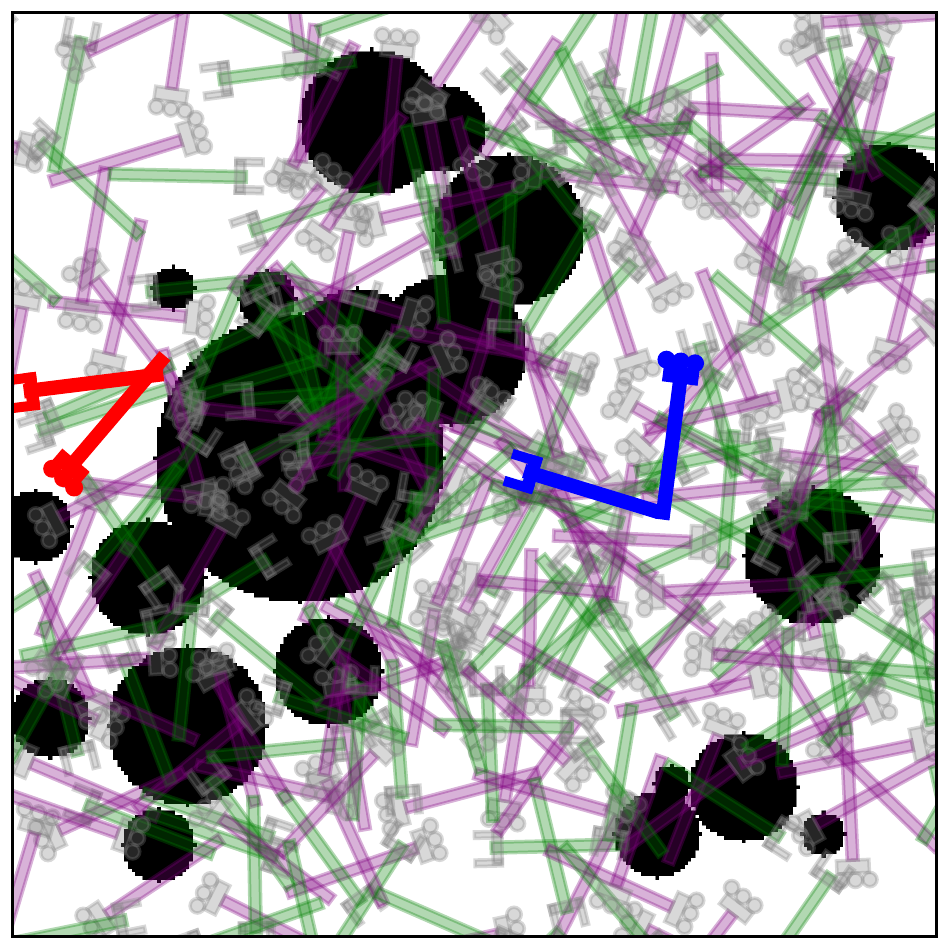}
            \caption{$\mathcal{U}(0,1)^d$
            \label{fig:samp-conditioning:uniform}
            }
        \end{subfigure}
    \caption{
         One hundred sampled configurations of $q\sim\SampleDist_\theta(q\given \cdots)$ in the 4D scenario
        with frozen weights.
         Blue and red robot arms represent the \textbf{\textcolor{blue}{initial}} and \textbf{\textcolor{red}{target}} state respectively, and the semi-transparent coloured robots arms (with colour-scheme that follows~\cref{fig:robotic-arm}) are the sampled configurations from $\SampleDist_\theta$;
         circular black blobs are obstacles.
         (a)--(d) are conditioned on the same latent variable and workspace information, except that~(\subref{fig:samp-conditioning:init-goal}) is also conditions on both the initial and target state,~(\subref{fig:samp-conditioning:init}) the initial state, and~(\subref{fig:samp-conditioning:goal}) the target state.
         For comparison purpose,~(\subref{fig:samp-conditioning:uniform}) is from a typical uniform distribution in traditional SBPs.
         When~(\subref{fig:samp-conditioning:init-goal}) $\SampleDist_\theta$ conditioned on states $q_\text{init}$, $q_\text{target}$ it resembles the region of optimal trajectory; 
         whereas when~(\subref{fig:samp-conditioning:none}) $\SampleDist_\theta$ is only conditioned on $\omega$, the distribution is less information specific, with wider deviation.
         Note that the projected robot arms overlap with obstacles because these are raw samples from the distribution without any collision checking with the simulator yet.
        \label{fig:samp-conditioning}
    }
\end{figure}

\vspace{-2mm}
\subsection{Planning information as conditional variables}\label{sec:Q-cond-incomplete-info}
\vspace{-1.5mm}

\Cref{fig:samp-conditioning} demonstrates sampling from the learned flow-based distribution $\SampleDist_\theta$ with different conditioning variables and a fixed latent vector.
Note that sampling is performed purely in \Cspace, but configurations are projected into the workspace for illustration purpose.
When provided with all available planning information (\cref{fig:samp-conditioning:init-goal}), $\SampleDist_\theta$ is able to capture a region that lies closely to the optimal trajectory.
Interestingly, \cref{fig:samp-conditioning:init,fig:samp-conditioning:goal} provides an insight into the effect of conditional variables with its asymmetric comparison.
Because $\qtarget$ (red robot arm) is at an area with limited visibility, sampling $\SampleDist_\theta$ without $\qtarget$ being conditioned naturally results in much less configurations around its nearby region at top-left corner (in \cref{fig:samp-conditioning:init}. Note that the same is true for \cref{fig:samp-conditioning:none}).
In contrast, $\qinit$ is at an open area; hence, even though $\qinit$ it not conditioned in~\cref{fig:samp-conditioning:goal}, its nearby region has considerably more sampled configurations.
This intuition is confirmed by~\cref{fig:samp-conditioning:none}, where the prior belief of $\SampleDist_\theta$ conditioned only on workspace information $\workspace$ closely resembles~\cref{fig:samp-conditioning:init} (although the distribution in~\cref{fig:samp-conditioning:none} is more spread out with higher variance due to no anchoring states).
For comparison, \cref{fig:samp-conditioning:uniform} illustrates the sampled configurations from a typical uniform sampler which samples everywhere but does not exploits any environment information.

\vspace{-2.5mm}
\section{Conclusion}
\vspace{-2mm}

We presented a novel sampling-based motion planner with conditional normalising flows that learns sampling strategies from contextual information.  
In all tested environments, our flow-based distribution $\SampleDist_\theta$ outperforms other methods by conditioning on all of the available planning information.
Even when provided with incomplete information, $\SampleDist_\theta$ is still able to efficiently sample from regions of high acceptability based on its prior experiences.
This property allows $\SampleDist_\theta$ to remain effective even in an explorative scenario where there exists no defined target state for the agent (similar to scenario in~\cref{fig:samp-conditioning:init} where only the robot's current state $\qinit$ is given).
PlannerFlows is computationally efficient, robust against mode-collapse, and provides informative conditional distributions for sampling.
The method is directly applicable to many existing SBPs, and experimental results show that it can dramatically improve the quality of samples as well as the overall runtime.

\clearpage

\printbibliography

\clearpage

\end{document}

%% file: _env_vars.tex
\definecolor{amethyst}{rgb}{0.6, 0.4, 0.8}
\definecolor{alizarin}{rgb}{0.82, 0.1, 0.26}
\definecolor{ashgrey}{rgb}{0.43, 0.5, 0.5}
\definecolor{yellow}{rgb}{1.0, 0.75, 0.0} %
\def\Cspace{\text{\emph{C-space}}\xspace}
\def\C{\ensuremath{C}\xspace}
\def\Cfree{\ensuremath{\C_\text{free}}\xspace}
\def\Cobs{\ensuremath{\C_\text{obs}}\xspace}
\def\Cobs{\ensuremath{\C_\text{obs}}\xspace}

\def\qinit{\ensuremath{q_\text{init}}\xspace}
\def\qtarget{\ensuremath{q_\text{target}}\xspace}

\newcommand*\workspace{\omega}

\makeatletter

\def\rrt{\textsc{rrt}}%
\newcommand*\RRT{%
    \@ifstar{%
        \rrt\textsuperscript{*}\xspace%
    }{%
        \rrt\xspace%
    }}

\def\rrdt{\textsc{rr}{\smaller d}\textsc{t}}%
\newcommand*\RRdT{%
    \@ifstar{%
        \rrdt\textsuperscript{*}\xspace%
    }{%
        \rrdt\xspace%
    }}

\def\prm{\textsc{prm}}%
\newcommand*\PRM{%
    \@ifstar{%
        \prm\textsuperscript{*}\xspace%
    }{%
        \prm\xspace%
    }}

\makeatother

\newcommand*\SampleDist{\mathcal{Q}}

\newcommand*\given[1][]{\:#1\vert\:}

\makeatletter
\newcommand{\overrightsmallarrow}{\mathpalette{\overarrowsmall@\rightarrowfill@}}
\newcommand{\overarrowsmall@}[3]{%
  \vbox{%
    \ialign{%
      ##\crcr
      #1{\smaller@style{#2}}\crcr
      \noalign{\nointerlineskip}%
      $\m@th\hfil#2#3\hfil$\crcr
    }%
  }%
}
\def\smaller@style#1{%
  \ifx#1\displaystyle\scriptstyle\else
    \ifx#1\textstyle\scriptstyle\else
      \scriptscriptstyle
    \fi
  \fi
}
\makeatother

\newcommand*\bLookOut[1]{%
    \ensuremath{\beta}\textsc{\footnotesize{-LookOut}}%
    \ifthenelse{\isempty{#1}}%
    {}{(#1)}%
}

\mathchardef\ordinarycolon\mathcode`\:
\mathcode`\:=\string"8000
\begingroup \catcode`\:=\active
  \gdef:{\mathrel{\mathop\ordinarycolon}}
\endgroup

%% file: expr-table.tex
\begin{tabular}{@{}clc@{}}
\toprule
\multicolumn{2}{c}{Planner}                                        & Total ($\times10^{3}$)  \\ \midrule
\multirow{8}{*}{2D} & \multicolumn{1}{l|}{RRT*}                    & $20.7\pm0.613$          \\
                    & \multicolumn{1}{l|}{RRT* flow}    & $\mathbf{14.4\pm0.251}$ \\
                    & \multicolumn{1}{l|}{Bi.RRT*}                 & $20.4\pm0.441$          \\
                    & \multicolumn{1}{l|}{Bi.RRT* flow} & $\mathbf{14.5\pm0.235}$ \\
                    & \multicolumn{1}{l|}{In.RRT*}                 & $19.7\pm2.23$           \\
                    & \multicolumn{1}{l|}{In.RRT* flow} & $18.7\pm3.69$           \\
                    & \multicolumn{1}{l|}{MPNet}                   & $20.5\pm0.303$          \\
                    & \multicolumn{1}{l|}{CVAE-based}              & $22.5\pm2.21$           \\ \midrule
\multirow{8}{*}{4D} & \multicolumn{1}{l|}{RRT*}                    & $61.8\pm13.8$           \\
                    & \multicolumn{1}{l|}{RRT* flow}    & $42.3\pm10.4$           \\
                    & \multicolumn{1}{l|}{Bi.RRT*}                 & $53.5\pm9.18$           \\
                    & \multicolumn{1}{l|}{Bi.RRT* flow} & $36.7\pm7.68$           \\
                    & \multicolumn{1}{l|}{In.RRT*}                 & $39.4\pm16.9$           \\
                    & \multicolumn{1}{l|}{In.RRT* flow} & $\mathbf{32.3\pm6.21}$  \\
                    & \multicolumn{1}{l|}{MPNet}                   & $60.4\pm11.6$           \\
                    & \multicolumn{1}{l|}{CVAE-based}              & $110\pm42.1$            \\ \midrule
\multirow{8}{*}{6D} & \multicolumn{1}{l|}{RRT*}                    & $36.0\pm12.8$           \\
                    & \multicolumn{1}{l|}{RRT* flow}    & $\mathbf{11.6\pm0.725}$ \\
                    & \multicolumn{1}{l|}{Bi.RRT*}                 & $33.9\pm11.3$           \\
                    & \multicolumn{1}{l|}{Bi.RRT* flow} & $\mathbf{11.5\pm0.658}$ \\
                    & \multicolumn{1}{l|}{In.RRT*}                 & $25.2\pm3.54$           \\
                    & \multicolumn{1}{l|}{In.RRT* flow} & $12.9\pm1.99$           \\
                    & \multicolumn{1}{l|}{MPNet}                   & $32.7\pm14.3$           \\
                    & \multicolumn{1}{l|}{CVAE-based}              & $17.3\pm0.927$          \\ \bottomrule
\end{tabular}